\documentclass[11pt,a4paper]{article}
\usepackage[hyperref]{emnlp2018}
\usepackage{times}
\usepackage{latexsym}
\usepackage{graphicx}
\usepackage{booktabs} 
\usepackage{multirow}
\usepackage{balance}
\usepackage{mathrsfs}
\usepackage{amsmath}
\usepackage{mathtools}
\usepackage{amsfonts}
\usepackage{enumitem}
\usepackage{url}

\usepackage{caption}
\usepackage{subcaption}
\usepackage{capt-of}

\aclfinalcopy 

\newcommand{\refeqn}[1]{Equation \ref{#1}}
\newcommand{\reffig}[1]{Figure \ref{#1}}
\newcommand{\reftbl}[1]{Table \ref{#1}}
\newcommand{\refsec}[1]{Section \ref{#1}}

\newcommand{\m}[1]{\mathcal{#1}}
\newcommand{\method}{RESIDE}
\newcommand{\lstm}{GRU}
\newcommand{\stepOne}{Syntactic Sentence Encoding}
\newcommand{\stepTwo}{Side Information Acquisition}
\newcommand{\stepThree}{Instance Set Aggregation}

\title{RESIDE: Improving Distantly-Supervised Neural Relation Extraction using Side Information}

\author{Shikhar Vashishth$^1$ \quad Rishabh Joshi$^2$ \thanks{\, This research was conducted during the author's internship at Indian Institute of Science.} \quad Sai Suman Prayaga$^1$ \quad \\ \textbf{Chiranjib Bhattacharyya}$^1$ \quad 	\textbf{Partha Talukdar}$^1$\\
	$^1$ Indian Institute of Science\\
	$^2$ Birla Institute of Technology and Science, Pilani \\
	{\tt \small \{shikhar,chiru,ppt\}@iisc.ac.in} \\
	{\tt \small f2014102@pilani.bits-pilani.ac.in, suman.sai14@gmail.com} \\
}

\date{}

\begin{document}
\maketitle
\begin{abstract}
Distantly-supervised Relation Extraction (RE) methods train an extractor by automatically aligning relation instances in a Knowledge Base (KB) with unstructured text. In addition to relation instances, KBs often contain other relevant side information, such as aliases of relations (e.g., \textit{founded} and \textit{co-founded} are aliases for the relation \textit{founderOfCompany}). RE models usually ignore such readily available side information. 
In this paper, we propose \method{}, a distantly-supervised neural relation extraction method which utilizes additional side information from KBs for improved relation extraction. It uses entity type and relation alias information for imposing soft constraints while predicting relations. 
\method{} employs Graph Convolution Networks (GCN) to encode syntactic information from text and improves performance even when limited side information is available. 
Through extensive experiments on benchmark datasets, we demonstrate \method{}'s effectiveness. We have made \method{}'s source code available to encourage reproducible research.
\end{abstract}

\section{Introduction}

\begin{figure*}[t]
	\centering
	\includegraphics[width=\textwidth]{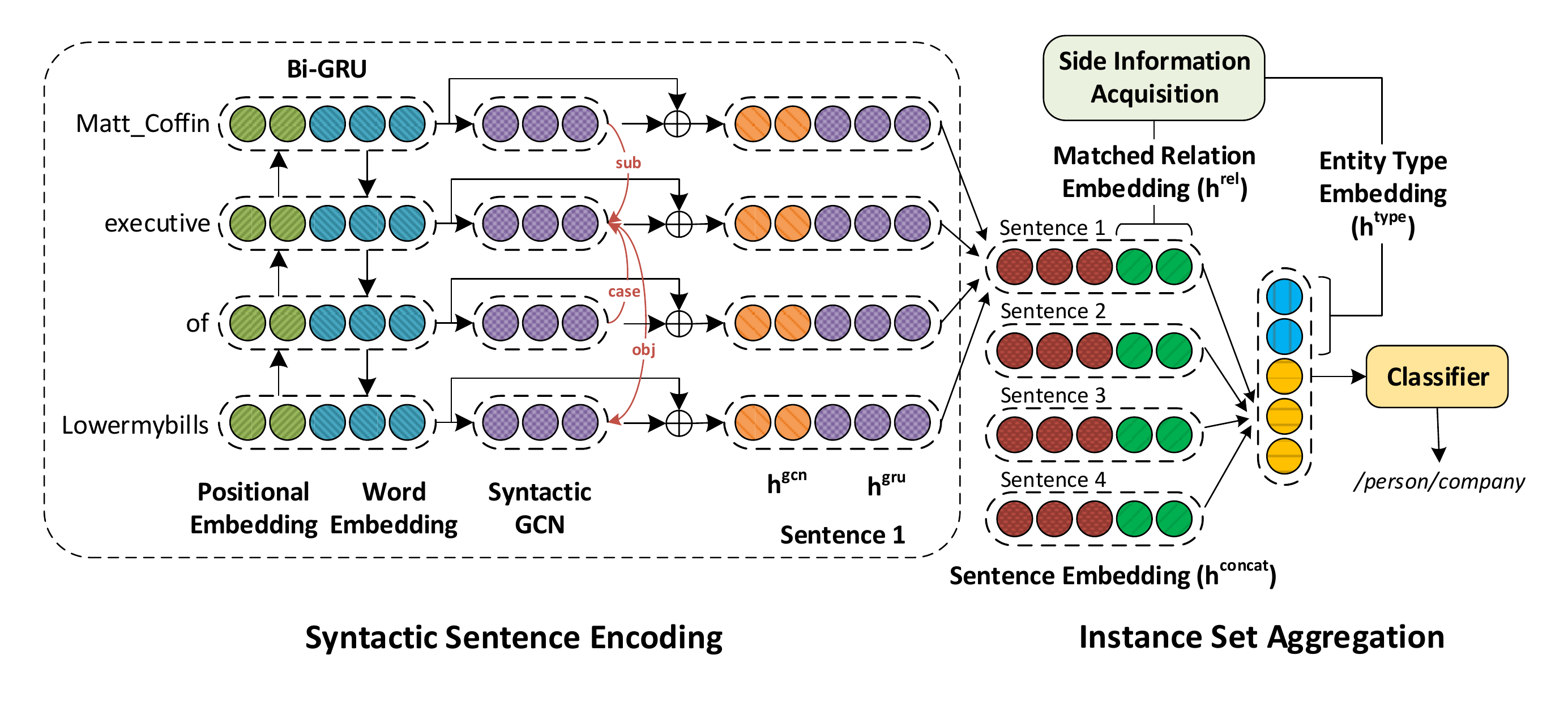}
	\caption{\label{fig:model_overview}Overview of \method{}. \method{} first encodes each sentence in the bag by concatenating embeddings (denoted by $\oplus$) from Bi-\lstm{} and Syntactic GCN for each token, followed by word attention. Then, sentence embedding is concatenated with relation alias information, which comes from the Side Information Acquisition Section (\reffig{fig:rel_alias}), before computing attention over sentences. Finally, bag representation with entity type information is fed to a softmax classifier. Please see \refsec{sec:details} for more details.}  
\end{figure*}

The construction of large-scale Knowledge Bases (KBs) like Freebase \cite{freebase_paper} and Wikidata \cite{wikidata_paper} has proven to be useful in many natural language processing (NLP) tasks like question-answering, web search, etc. However, these KBs are not exhaustive. Relation Extraction (RE) attempts to fill this gap by extracting semantic relationships between entity pairs from plain text. This task can be modeled as a simple classification problem after the entity pairs are specified. Formally, given an entity pair ($e_1$,$e_2$) from the KB and an entity annotated sentence (or instance), we aim to predict the relation $r$, from a predefined relation set, that exists between $e_1$ and $e_2$. If no relation exists, we simply label it \textit{NA}.

Most supervised relation extraction methods require large labeled training data which is expensive to construct. Distant Supervision (DS) \cite{mintz2009distant} helps with the construction of this dataset automatically, under the assumption that if two entities have a relationship in a KB, then all sentences mentioning those entities express the same relation. While this approach works well in generating large amounts of training instances, the DS assumption does not hold in all cases. \citet{riedel2010modeling,hoffmann2011knowledge,surdeanu2012multi} propose multi-instance based learning to relax this assumption. However, they use NLP tools to extract features, which can be noisy. 


Recently, neural models have demonstrated promising performance on RE. \citet{zeng2014relation,zeng2015distant} employ Convolutional Neural Networks (CNN) to learn representations of instances. For alleviating noise in distant supervised datasets, attention has been utilized by \cite{lin2016neural,bgwa_paper}.
Syntactic information from dependency parses has been used by \cite{mintz2009distant,see_paper} for capturing long-range dependencies between tokens. Recently proposed Graph Convolution Networks (GCN) \cite{Defferrard:2016:CNN:3157382.3157527} have been effectively employed for encoding this information \cite{gcn_srl,gcn_nmt}. However, all the above models rely only on the noisy instances from distant supervision for RE.

Relevant side information can be effective for improving RE. For instance, in the sentence, \textit{Microsoft was started by Bill Gates.}, the type information of \textit{Bill Gates (person)}  and \textit{Microsoft (organization)} can be helpful in predicting the correct relation \textit{founderOfCompany}. This is because every relation constrains the type of its target entities. 
Similarly, relation phrase \textit{``was started by"} extracted using Open Information Extraction (Open IE) methods can be useful, given that the aliases of relation \textit{founderOfCompany}, e.g., \textit{founded, co-founded, etc.}, are available. KBs used for DS readily provide such information which has not been completely exploited by current models.


In this paper, we propose \method{}, a novel distant supervised relation extraction method which utilizes additional supervision from KB through its neural network based architecture.
\method{} makes principled use of entity type and relation alias information from KBs, to impose soft constraints while predicting the relation. 
It uses encoded syntactic information obtained from Graph Convolution Networks (GCN), along with embedded side information, to improve neural relation extraction.
Our contributions can be summarized as follows:
\begin{itemize}[itemsep=2pt,parsep=0pt,partopsep=0pt,leftmargin=*]
	\item We propose \method{}, a novel neural method which utilizes additional supervision from KB in a principled manner for improving distant supervised RE.
	\item \method{} uses Graph Convolution Networks (GCN) for modeling syntactic information and has been shown to perform competitively even with limited side information.
	\item Through extensive experiments on benchmark datasets, we demonstrate \method{}'s effectiveness over state-of-the-art baselines.
\end{itemize} 
\method{}'s source code and datasets used in the paper are available at \url{http://github.com/malllabiisc/RESIDE}.
\section{Related Work}
\label{sec:related_work}

\textbf{Distant supervision:} Relation extraction is the task of identifying the relationship between two entity mentions in a sentence. In supervised paradigm, the task is considered as a multi-class classification problem but suffers from lack of large labeled training data. To address this limitation, \cite{mintz2009distant} propose distant supervision (DS) assumption for creating large datasets, by heuristically aligning text to a given Knowledge Base (KB). As this assumption does not always hold true, some of the sentences might be wrongly labeled. To alleviate this shortcoming, \citet{riedel2010modeling} relax distant supervision for multi-instance single-label learning. Subsequently, for handling overlapping relations between entities \cite{hoffmann2011knowledge,surdeanu2012multi} propose multi-instance multi-label learning paradigm. 

\textbf{Neural Relation Extraction:} The performance of the above methods strongly rely on the quality of hand engineered features. \citet{zeng2014relation} propose an end-to-end CNN based method which could automatically capture relevant lexical and sentence level features. This method is further improved through piecewise max-pooling by \cite{zeng2015distant}.
\citet{lin2016neural,candis_paper} use attention \cite{bahdanau+al-2014-nmt} for learning from multiple valid sentences. We also make use of attention for learning sentence and bag representations. 

Dependency tree based features have been found to be relevant for relation extraction \cite{mintz2009distant}. \citet{see_paper} use them for getting promising results through a recursive tree-GRU based model. In \method{}, we make use of recently proposed Graph Convolution Networks \cite{Defferrard:2016:CNN:3157382.3157527,kipf2016semi}, which have been found to be quite effective for modelling syntactic information \cite{gcn_srl,gcn_event,neuraldater_paper}.

\textbf{Side Information in RE:} Entity description from KB has been utilized for RE \cite{entity_description}, but such information is not available for all entities. Type information of entities has been used by \citet{figer_paper,Liu2014ExploringFE} as features in their model. \citet{typeinfo2017} also attempt to mitigate noise in DS through their joint entity typing and relation extraction model. However, KBs like Freebase readily provide reliable type information which could be directly utilized. In our work, we make principled use of entity type and relation alias information obtained from KB. We also use unsupervised Open Information Extraction (Open IE) methods \cite{ollie,stanford_openie}, which automatically discover possible relations without the need of any predefined ontology, which is used as a side information as defined in \refsec{sec:sideinfo}.



\section{Background: Graph Convolution Networks (GCN)}
\label{sec:gcn_background}

In this section, we provide a brief overview of Graph Convolution Networks (GCN) for graphs with directed and labeled edges, as used in \cite{gcn_srl}. 

\subsection{GCN on Labeled Directed Graph}

For a directed graph, $\m{G} = (\m{V}, \m{E})$, where $\m{V}$ and $\m{E}$ represent the set of vertices and edges respectively, an edge from node $u$ to node $v$ with label $l_{uv}$ is represented as $(u, v, l_{uv})$. Since, information in directed edge does not necessarily propagate along its direction, following \cite{gcn_srl} we define an updated edge set $\m{E'}$ which includes inverse edges $(v,u,l_{uv}^{-1})$ and self-loops $(u,u,\top)$ along with the original edge set $\m{E}$, where $\top$ is a special symbol to denote self-loops. For each node $v$ in $\m{G}$, we have an initial representation $x_{v} \in \mathbb{R}^{d}\text{, }\forall v \in \m{V}$. On employing GCN, we get an updated $d$-dimensional hidden representation $h_v \in \mathbb{R}^{d}\text{, }\forall v \in \m{V}$, by considering only its immediate neighbors \cite{kipf2016semi}. This can be formulated as:
\begin{equation*}
h_{v} = f \left(\sum_{u \in \m{N}(v)}\left(W_{l_{uv}}x_{u} + b_{l_{uv}}\right)\right).
\end{equation*}
Here, $W_{l_{uv}} \in \mathbb{R}^{d \times d}$ and $b_{l_{uv}} \in \mathbb{R}^{d}$ are label dependent model parameters which are trained based on the downstream task. $\m{N}(v)$ refers to the set of neighbors of $v$ based on $\m{E'}$ and $f$ is any non-linear activation function. In order to capture multi-hop neighborhood, multiple GCN layers can be stacked. Hidden representation of node $v$ in this case after $k^{th}$ GCN layer is given as:
\begin{equation*}
h_{v}^{k+1} = f \left(\sum_{u \in \m{N}(v)}\left(W^{k}_{l_{uv}}h_{u}^{k} + b^{k}_{l_{uv}}\right)\right).
\end{equation*}
\subsection{Integrating Edge Importance} 
In automatically constructed graphs, some edges might be erroneous and hence need to be discarded. Edgewise gating in GCN by \cite{gcn_nmt,gcn_srl} allows us to alleviate this problem by subduing the noisy edges. This is achieved by assigning a relevance score to each edge in the graph. At $k^{th}$ layer, the importance of an edge $(u, v, l_{uv})$ is computed as:
\begin{equation}
\label{eqn:edge_gating}
g^{k}_{uv} = \sigma \left( h^{k}_u \cdot \hat{w}^{k}_{l_{uv}}  + \hat{b}^{k}_{l_{uv}} \right),
\end{equation}
Here, $\hat{w}^{k}_{l_{uv}} \in \mathbb{R}^{m}$ and $\hat{b}^{k}_{l_{uv}} \in \mathbb{R}$ are parameters which are trained and $\sigma(\cdot)$ is the sigmoid function. With edgewise gating, the final GCN embedding for a node $v$ after $k^{th}$ layer is given as:
\begin{equation}
\label{eqn:gcn_directed}
h^{k+1}_{v} = f\left(\sum_{u \in \m{N}(v)} g^{k}_{uv} \times \left({W}^{k}_{l_{uv}} h^{k}_{u} + b^{k}_{l_{uv}}\right)\right).
\end{equation}


\section{\method{} Overview}
\label{sec:overview}

In multi-instance learning paradigm, we are given a bag of sentences (or instances) $\{s_1, s_2, ... s_n\}$ for a given entity pair, the task is to predict the relation between them.
\method{} consists of three components for learning a representation of a given bag, which is fed to a softmax classifier. We briefly present the components of \method{} below. Each component will be described in detail in the subsequent sections. The overall architecture of \method{} is shown in \reffig{fig:model_overview}.

\begin{enumerate}[itemsep=2pt,parsep=0pt,partopsep=0pt,leftmargin=*]
	\item \textbf{\stepOne{}:} \method{} uses a Bi-\lstm{} over the concatenated positional and word embedding for encoding the local context of each token. For capturing long-range dependencies, GCN over dependency tree is employed and its encoding is appended to the representation of each token. Finally, attention over tokens is used to subdue irrelevant tokens and get an embedding for the entire sentence. More details in \refsec{sec:sent_encoder}.
	
	\item \textbf{\stepTwo{}:} In this module, we use additional supervision from KBs and utilize Open IE methods for getting relevant side information. This information is later utilized by the model as described in \refsec{sec:sideinfo}.
	
	\item \textbf{\stepThree{}:} In this part, sentence representation from syntactic sentence encoder is concatenated with the \textit{matched relation embedding} obtained from the previous step. Then, using attention over sentences, a representation for the entire bag is learned. This is then concatenated with \textit{entity type embedding} before feeding into the softmax classifier for relation prediction. Please refer to \refsec{sec:rel_ext} for more details.
	
\end{enumerate}
\section{\method{} Details}
\label{sec:details}

In this section, we provide the detailed description of the components of \method{}.

\begin{figure*}[t]
	\centering
	\includegraphics[width=\textwidth]{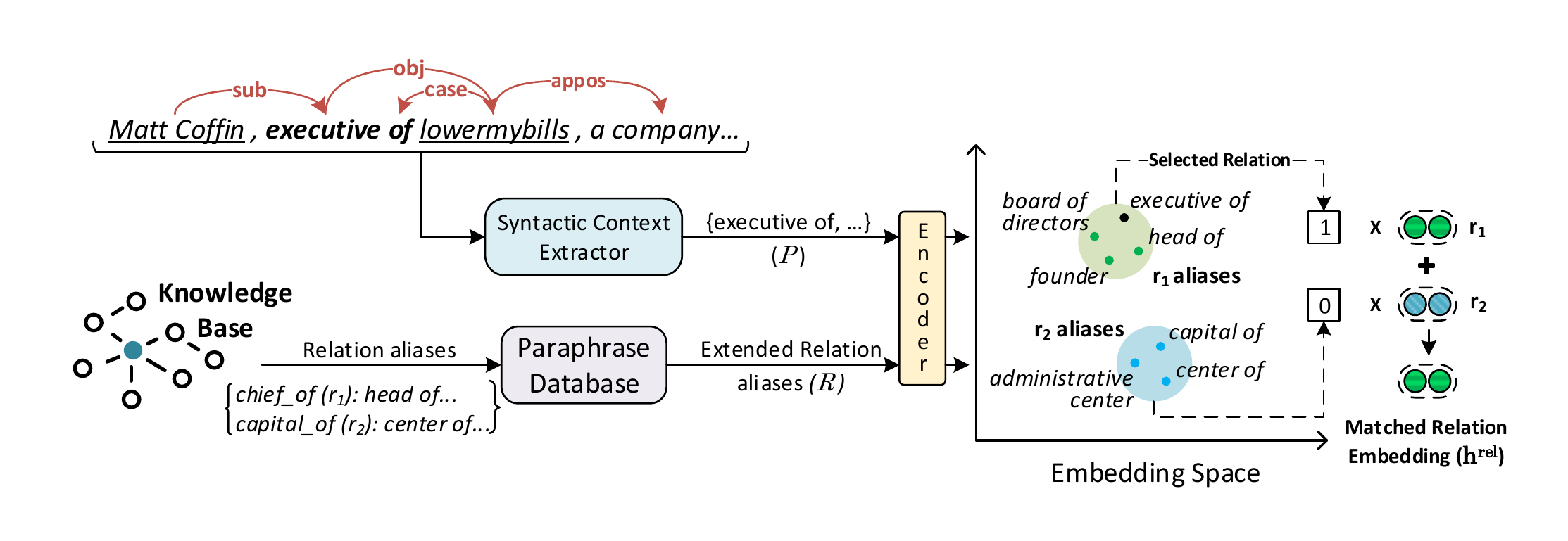}
	\caption{\label{fig:rel_alias}
	Relation alias side information extraction for a given sentence. First, Syntactic Context Extractor identifies relevant relation phrases $\m{P}$ between target entities. They are then matched in the embedding space with the extended set of relation aliases $\m{R}$ from KB. Finally, the relation embedding corresponding to the closest alias is taken as relation alias information. Please refer \refsec{sec:sideinfo}. } 
\end{figure*}

\subsection{\stepOne{}}
\label{sec:sent_encoder}
For each sentence in the bag $s_i$ with $m$ tokens $\{w_1, w_2, ... w_m\}$, we first represent each token by $k$-dimensional GloVe embedding \cite{pennington2014glove}. For incorporating relative position of tokens with respect to target entities, we use $p$-dimensional position embeddings, as used by \cite{zeng2014relation}. The combined token embeddings are stacked together to get the sentence representation $\m{H} \in \mathbb{R}^{m \times (k+2p)}$. Then, using Bi-\lstm{} \cite{gru_paper} over $\m{H}$, we get the new sentence representation $\m{H}^{gru} \in \mathbb{R}^{m \times d_{gru}}$, where $d_{gru}$ is the hidden state dimension. Bi-\lstm{}s have been found to be quite effective in encoding the context of tokens in several tasks \cite{Sutskever:2014:SSL:2969033.2969173,rnn_speech_recog}.

Although Bi-\lstm{} is capable of capturing local context, it fails to capture long-range dependencies which can be captured through dependency edges. Prior works \cite{mintz2009distant,see_paper} have exploited features from syntactic dependency trees for improving relation extraction. Motivated by their work, we employ Syntactic Graph Convolution Networks for encoding this information. For a given sentence, we generate its dependency tree using Stanford CoreNLP \cite{stanford_corenlp}. We then run GCN over the dependency graph and use \refeqn{eqn:gcn_directed} for updating the embeddings, taking $\m{H}^{gru}$ as the input. Since dependency graph has 55 different edge labels, incorporating all of them over-parameterizes the model significantly. Therefore, following \cite{gcn_srl,gcn_event,neuraldater_paper} we use only three edge labels based on the direction of the edge \{\textit{forward ($\rightarrow$), backward ($\leftarrow$), self-loop ($\top$)}\}. We define the new edge label $L_{uv}$ for an edge $(u,v,l_{uv})$ as follows:
\[ 
	L_{uv} =
	\begin{cases} 
		\rightarrow & \text{if edge exists in dependency parse} \\
		\leftarrow 	& \text{if edge is an inverse edge}	\\
		\top 		& \text{if edge is a self-loop}
	\end{cases}
\]

For each token $w_i$, GCN embedding $h^{gcn}_{i_{k+1}} \in \mathbb{R}^{d_{gcn}}$ after $k^{th}$ layer is defined as:
\[
h^{gcn}_{i_{k+1}} = f \Bigg(\sum_{u \in \m{N}(i)} g_{iu}^{k} \times \left(W_{L_{iu}}^{k}h^{gcn}_{u_{k}} + b_{L_{iu}}^{k}\right) \Bigg).
\]
Here, $g_{iu}^{k}$ denotes edgewise gating as defined in \refeqn{eqn:edge_gating} and $L_{iu}$ refers to the edge label defined above. We use ReLU as activation function $f$, throughout our experiments. The syntactic graph encoding from GCN is appended to Bi-\lstm{} output to get the final token representation, $h^{concat}_i$ as $[h^{gru}_i; h^{gcn}_{i^{k+1}}]$.
Since, not all tokens are equally relevant for RE task, we calculate the degree of relevance of each token using attention as used in \cite{bgwa_paper}. For token $w_i$ in the sentence, attention weight $\alpha_{i}$ is calculated as:
$$ \alpha_{i} = \dfrac{\text{exp}(u_{i})}{\sum_{j=1}^{m}{\text{exp}(u_{j})}} \, \, \text{ where, } u_{i} = h^{concat}_{i} \cdot r .$$
where $r$ is a random query vector and $u_{i}$ is the relevance score assigned to each token. Attention values $\{\alpha_{i}\}$ are calculated by taking softmax over $\{u_{i}\}$. The representation of a sentence is given as a weighted sum of its tokens,
$ s = \sum_{j=1}^{m}{\alpha_{i}h^{concat}_i}$.

\subsection{\stepTwo{}}
\label{sec:sideinfo}

Relevant side information has been found to improve performance on several tasks \cite{figer_paper,cesi_paper}. In distant supervision based relation extraction, since the entities are from a KB, knowledge about them can be utilized to improve relation extraction. Moreover, several unsupervised relation extraction methods (Open IE) \cite{stanford_openie,ollie} allow extracting relation phrases between target entities without any predefined ontology and thus can be used to obtain relevant side information. In \method{}, we employ Open IE methods and additional supervision from KB for improving neural relation extraction. 

\subsubsection*{Relation Alias Side Information}
\label{sec:alias_sideinfo}

\method{} uses Stanford Open IE \cite{stanford_openie} for extracting relation phrases between target entities, which we denote by $\m{P}$. As shown in \reffig{fig:rel_alias}, for the sentence \textit{Matt Coffin, executive of lowermybills, a company..}, Open IE methods extract \textit{``executive of"} between \textit{Matt Coffin} and \textit{lowermybills}. Further, we extend $\m{P}$ by including tokens at one hop distance in dependency path from target entities. Such features from dependency parse have been exploited in the past by \cite{mintz2009distant,see_paper}. The degree of match between the extracted phrases in $\m{P}$ and aliases of a relation can give important clues about the relevance of that relation for the sentence. 
Several KBs like Wikidata provide such relation aliases, which can be readily exploited. In \method{}, we further expand the relation alias set using Paraphrase database (PPDB) \cite{ppdb2_paper}. We note that even for cases when aliases for relations are not available, providing only the names of relations give competitive performance. We shall explore this point further in  \refsec{sec:results_rel_side}.

For matching $\m{P}$ with the PPDB expanded relation alias set $\m{R}$, we project both in a $d$-dimensional space using GloVe embeddings \cite{pennington2014glove}. Projecting phrases using word embeddings helps to further expand these sets, as semantically similar words are closer in embedding space \cite{word2vec_paper,pennington2014glove}. Then, for each phrase $p \in \m{P}$, we calculate its cosine distance from all relation aliases in $\m{R}$ and take the relation corresponding to the closest relation alias as a matched relation for the sentence. We use a threshold on cosine distance to remove noisy aliases. In \method{}, we define a $k_r$-dimensional embedding for each relation which we call as \textit{matched relation embedding} $(h^{rel})$. For a given sentence, $h^{rel}$ is concatenated with its representation $s$, obtained from syntactic sentence encoder (\refsec{sec:sent_encoder}) as shown in \reffig{fig:model_overview}. For sentences with $|\m{P}|>1$, we might get multiple matched relations. In such cases, we take the average of their embeddings. We hypothesize that this helps in improving the performance and find it to be true as shown in \refsec{sec:results}. 

\subsubsection*{Entity Type Side Information}
\label{sec:type_sideinfo}
Type information of target entities has been shown to give promising results on relation extraction \cite{figer_paper,typeinfo2017}. Every relation puts some constraint on the type of entities which can be its subject and object. For example, the relation \textit{person/place\_of\_birth} can only occur between a \textit{person} and a \textit{location}. Sentences in distance supervision are based on entities in KBs, where the type information is readily available.

In \method{}, we use types defined by FIGER \cite{figer_paper} for entities in Freebase. For each type, we define a $k_t$-dimensional embedding which we call as \textit{entity type embedding} ($h^{type}$). For cases when an entity has multiple types in different contexts, for instance, \textit{Paris} may have types \textit{government} and \textit{location}, we take the average over the embeddings of each type. We concatenate the \textit{entity type embedding} of target entities to the final bag representation before using it for relation classification. To avoid over-parameterization, instead of using all fine-grained 112 entity types, we use 38 coarse types which form the first hierarchy of FIGER types.

\subsection{\stepThree{}}
\label{sec:rel_ext}

For utilizing all valid sentences, following \cite{lin2016neural,bgwa_paper}, we use attention over sentences to obtain a representation for the entire bag. Instead of directly using the sentence representation $s_i$ from \refsec{sec:sent_encoder}, we concatenate the embedding of each sentence with \textit{matched relation embedding} $h_i^{rel}$ as obtained from \refsec{sec:sideinfo}. 
The attention score $\alpha_i$ for $i^{th}$ sentence is formulated as:
$$ \alpha_{i} = \dfrac{\text{exp}(\hat{s}_i \cdot q)}{\sum_{j=1}^{n}{\text{exp}(\hat{s}_j \cdot q)}} \, \, \text{ where, } \hat{s}_{i} = [s_{i};h^{rel}_i] .$$

here $q$ denotes a random query vector. The bag representation $\m{B}$, which is the weighted sum of its sentences, is then concatenated with the \textit{entity type embeddings} of the subject ($h^{type}_{sub}$) and object ($h^{type}_{obj}$) from \refsec{sec:sideinfo} to obtain $\hat{\m{B}}$.
$$ \hat{\m{B}} = [\m{B};h^{type}_{sub}; h^{type}_{obj}]  \, \, \text{ where, } \m{B} = \sum_{i=1}^{n}{\alpha_i \hat{s}_i}.$$
Finally, $\hat{\m{B}}$ is fed to a softmax classifier to get the probability distribution over the relations.
$$p(y) = \mathrm{Softmax}(W \cdot \hat{\m{B}} + b).$$




\begin{table}[t]
	\small
	\begin{tabular}{ccccc}
		\toprule
		Datasets 	& Split & \# Sentences 	& \# Entity-pairs \\
		\midrule
		\multirow{3}{*}{\shortstack{Riedel\\(\# Relations: 53)}} & Train & 455,771 & 233,064 \\
		 & Valid & 114,317 & 58,635 \\
		 & Test  & 172,448 & 96,678 \\
		 \midrule
 		\multirow{3}{*}{\shortstack{GDS\\(\# Relations: 5)}} & Train & 11,297 & 6,498 \\
 		& Valid & 1,864 & 1,082 \\
 		& Test  & 5,663 & 3,247 \\
		\bottomrule
		\addlinespace
	\end{tabular}
	\caption{\label{tb:datasets}Details of datasets used. Please see \refsec{sec:datasets} for more details. }
\end{table}

\section{Experimental Setup}
\subsection{Datasets}
\label{sec:datasets}

In our experiments, we evaluate the models on Riedel and Google Distant Supervision (GDS) dataset. Statistics of the datasets is summarized in \reftbl{tb:datasets}. Below we described each in detail\footnote{Data splits and hyperparameters are in supplementary.}.
\begin{enumerate}[leftmargin=*]
	\item \textbf{Riedel:} The dataset is developed by \cite{riedel2010modeling} by aligning Freebase relations with New York Times (NYT) corpus, where sentences from the year 2005-2006 are used for creating the training set and from the year 2007 for the test set. The entity mentions are annotated using Stanford NER \cite{finkel2005incorporating} and are linked to Freebase. The dataset has been widely used for RE by \cite{hoffmann2011knowledge,surdeanu2012multi} and more recently by \cite{lin2016neural,feng2017effective,see_paper}. 
	
	\item \textbf{GIDS:} \citet{bgwa_paper} created Google IISc Distant Supervision (GIDS) dataset by extending the Google relation extraction corpus\footnote{\href{https://research.googleblog.com/2013/04/50000-lessons-on-how-to-read-relation.html}{https://research.googleblog.com/2013/04/50000-lessons-on-how-to-read-relation.html}} with additional instances for each entity pair. The dataset assures that the at-least-one assumption of multi-instance learning, holds. This makes automatic evaluation more reliable and thus removes the need for manual verification.
\end{enumerate}


\begin{figure*}[t]
	\begin{minipage}{\textwidth}
		\captionsetup{type=figure} 
		\centering
		\subcaptionbox{Riedel dataset}
		{\includegraphics[width=.495\textwidth]{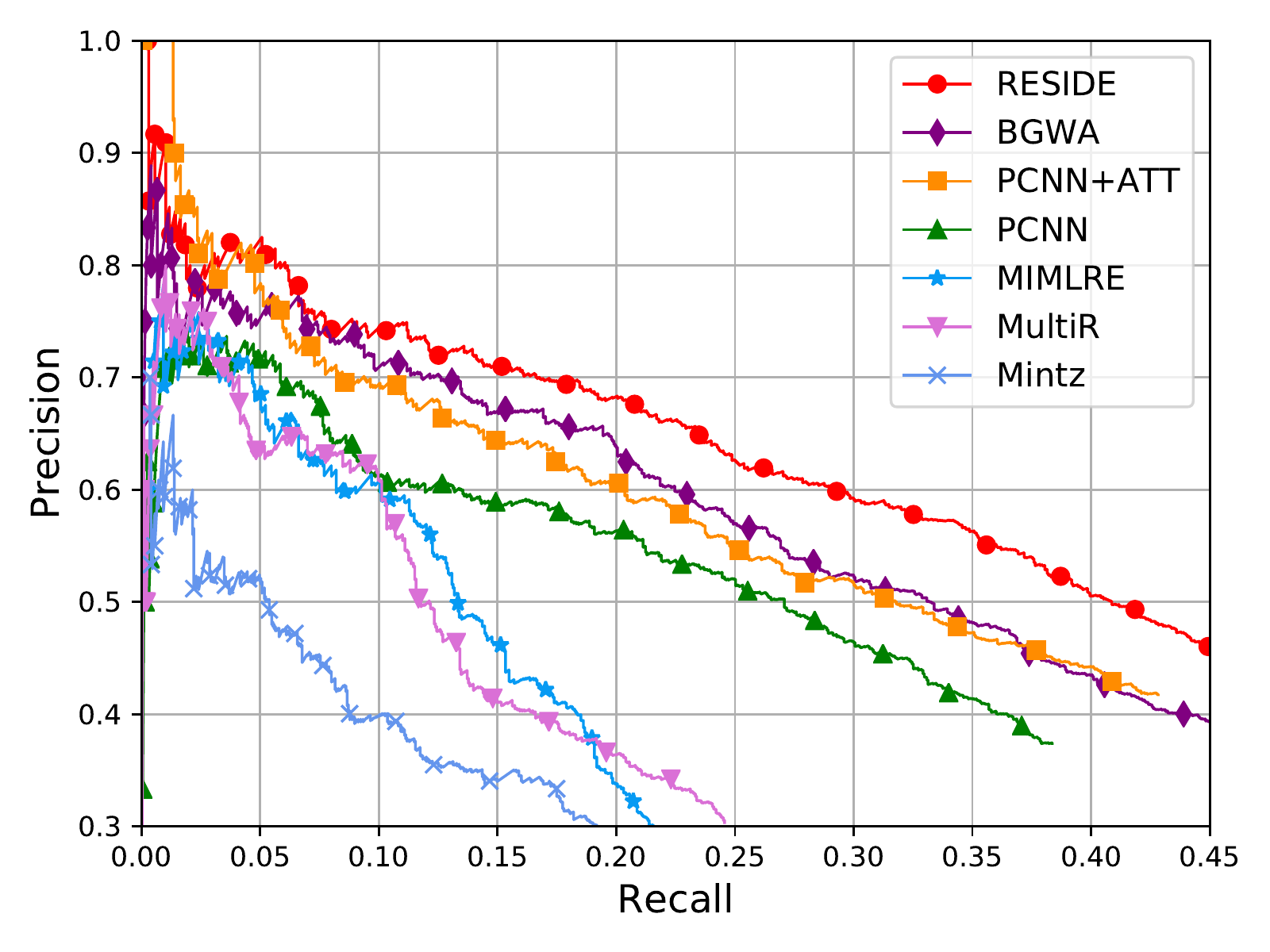}}
		\subcaptionbox{GIDS dataset}
		{\includegraphics[width=.495\textwidth]{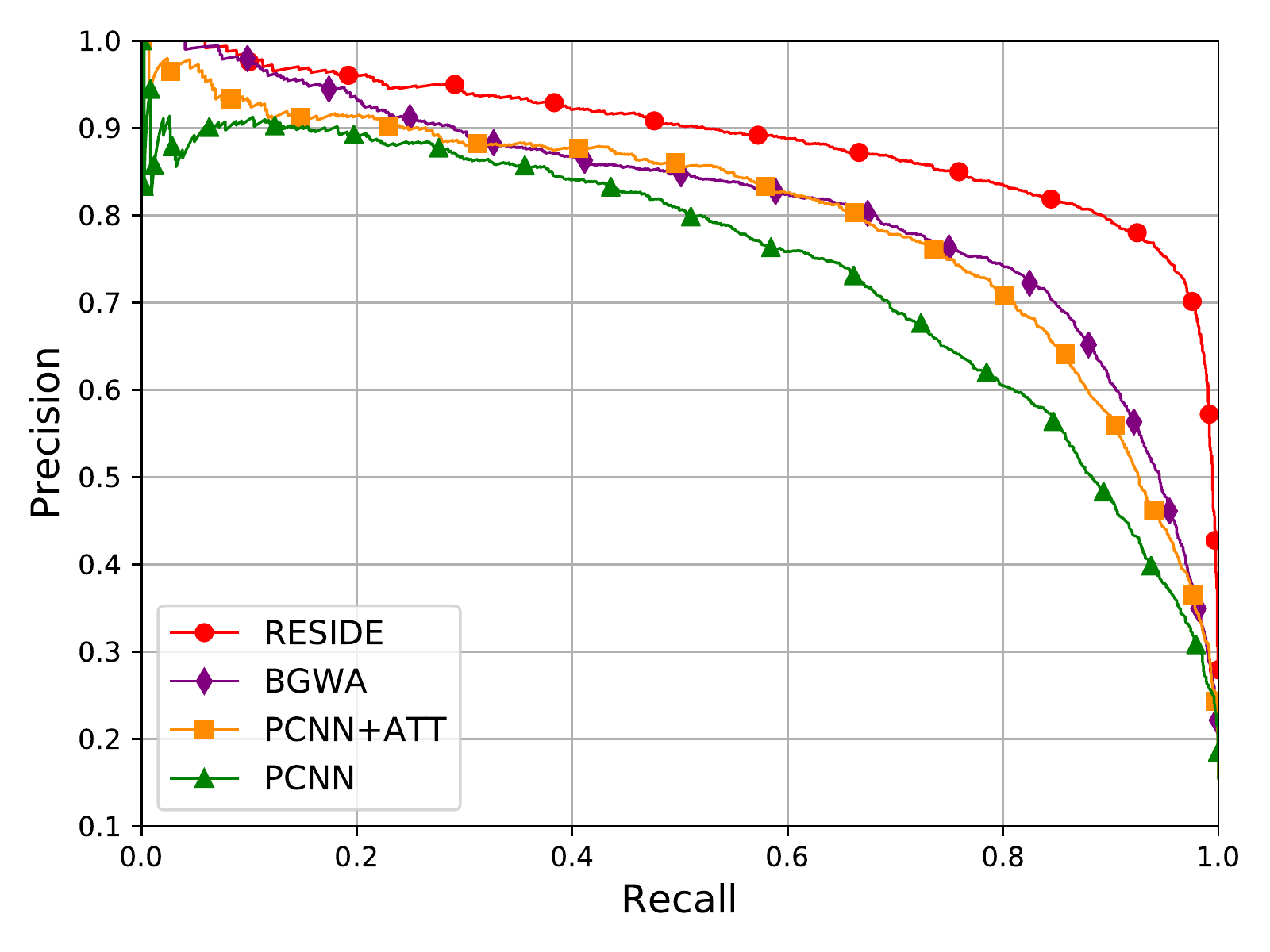}}
		\caption[main]{\label{fig:main_pr}Comparison of Precision-recall curve. \method{} achieves higher precision over the entire range of recall than all the baselines on both datasets. Please refer \refsec{sec:results_main} for more details.}
	\end{minipage}
\end{figure*}

\subsection{Baselines}
\label{sec:baselines}
For evaluating \method{}, we compare against the following baselines:

\begin{itemize}[itemsep=2pt,parsep=0pt,partopsep=0pt,leftmargin=*]
	\item \textbf{Mintz:} Multi-class logistic regression model proposed by \cite{mintz2009distant} for distant supervision paradigm.
	\item \textbf{MultiR:} Probabilistic graphical model for multi instance learning by \cite{hoffmann2011knowledge} 
	\item \textbf{MIMLRE:} A graphical model which jointly models multiple instances and multiple labels. More details in \cite{surdeanu2012multi}.
	\item \textbf{PCNN:} A CNN based relation extraction model by \cite{zeng2015distant} which uses piecewise max-pooling for sentence representation.
	\item \textbf{PCNN+ATT:} A piecewise max-pooling over CNN based model which is used by \cite{lin2016neural} to get sentence representation followed by attention over sentences.
	\item \textbf{BGWA:} Bi-GRU based relation extraction model with word and sentence level attention \cite{bgwa_paper}.
	\item \textbf{\method{}:} The method proposed in this paper, please refer \refsec{sec:details} for more details.
\end{itemize}

\subsection{Evaluation Criteria}
Following the prior works \cite{lin2016neural,feng2017effective}, we evaluate the models using held-out evaluation scheme. This is done by comparing the relations discovered from test articles with those in Freebase. We evaluate the performance of models with Precision-Recall curve and top-N precision (P@N) metric in our experiments.


\section{Results}
\label{sec:results}

In this section we attempt to answer the following questions:
\begin{itemize}[itemsep=2pt,topsep=4pt,parsep=0pt,partopsep=0pt]
	\item[Q1.] Is \method{} more effective than existing approaches for distant supervised RE? (\ref{sec:results_main})
	\item[Q2.] What is the effect of ablating different components on \method{}'s performance? (\ref{sec:results_sideinfo})
	\item[Q3.] How is the performance affected in the absence of relation alias information? (\ref{sec:results_rel_side})
\end{itemize}

\begin{table*}[t!]
	\centering
	\begin{small}
		\begin{tabular}{lccc|ccc|ccc}
			\toprule
			& \multicolumn{3}{c}{One} & \multicolumn{3}{c}{Two} & \multicolumn{3}{c}{All}\\ 
			\cmidrule(r){2-4} \cmidrule(r){5-7} \cmidrule(r){8-10} 
			& P@100 & P@200 & P@300 & P@100 & P@200 & P@300 & P@100 & P@200 & P@300 \\
			\midrule
			PCNN		& 73.3	& 64.8	& 56.8	& 70.3	& 67.2	& 63.1	& 72.3	& 69.7	& 64.1 \\ 
			PCNN+ATT	& 73.3	& 69.2	& 60.8	& 77.2	& 71.6	& 66.1	& 76.2	& 73.1	& 67.4 \\
			BGWA		& 78.0	& 71.0	& 63.3	& 81.0	& 73.0	& 64.0	& 82.0	& 75.0	& 72.0 \\
			\method{}	& \textbf{80.0}	& \textbf{75.5}	& \textbf{69.3}	& \textbf{83.0}	& \textbf{73.5}	& \textbf{70.6}	& \textbf{84.0}	& \textbf{78.5}	& \textbf{75.6} \\
			\bottomrule
			\addlinespace
		\end{tabular}
		\caption{\label{tb:np_canonicalization}P@N for relation extraction using variable number of sentences in bags (with more than one sentence) in Riedel dataset. Here, One, Two and All represents the number of sentences randomly selected from a bag. \method{} attains improved precision in all settings. More details in \refsec{sec:results_main}}
	\end{small}
\end{table*}

%


\subsection{Performance Comparison}
\label{sec:results_main}
For evaluating the effectiveness of our proposed method, \method{}, we compare it against the baselines stated in \refsec{sec:baselines}. We use only the neural baselines on GDS dataset. The Precision-Recall curves on Riedel and GDS are presented in \reffig{fig:main_pr}. Overall, we find that \method{} achieves higher precision over the entire recall range on both the datasets. All the non-neural baselines could not perform well as the features used by them are mostly derived from NLP tools which can be erroneous. RESIDE outperforms PCNN+ATT and BGWA which indicates that incorporating side information helps in improving the performance of the model. The higher performance of BGWA and PCNN+ATT over PCNN shows that attention helps in distant supervised RE. Following \cite{lin2016neural,softlabel_paper}, we also evaluate our method with different number of sentences. Results summarized in \reftbl{tb:np_canonicalization}, show the improved precision of \method{} in all test settings, as compared to the neural baselines, which demonstrates the efficacy of our model.


\subsection{Ablation Results}
\label{sec:results_sideinfo}
\begin{figure}[t]
	\centering
	\includegraphics[width=\columnwidth]{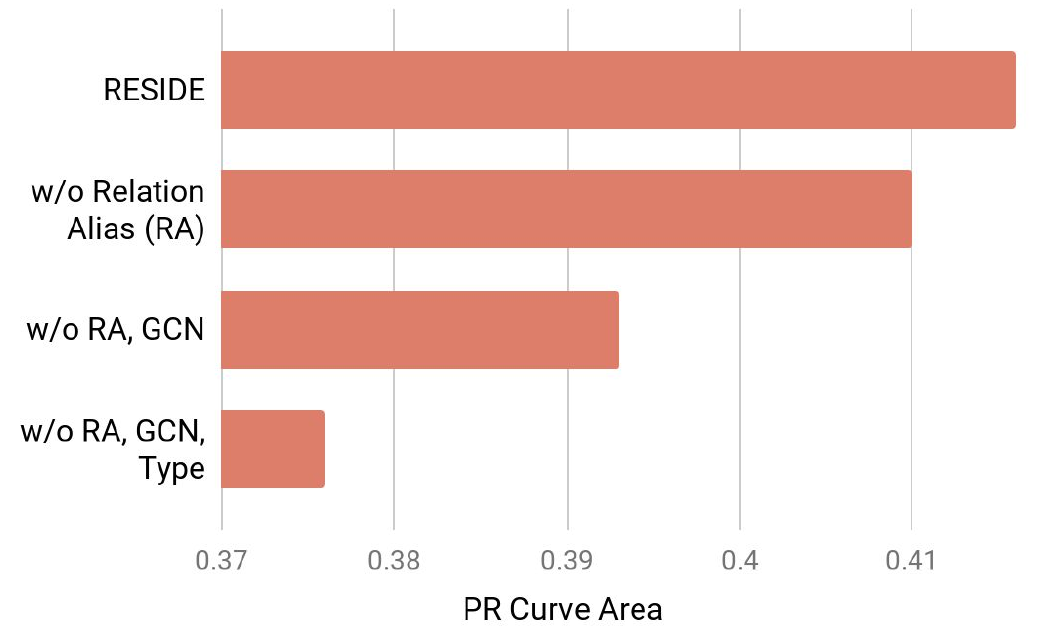}
	\caption{\label{fig:ablation}Performance comparison of different ablated version of \method{} on Riedel dataset. Overall, GCN and side information helps \method{} improve performance. Refer \refsec{sec:results_sideinfo}.}
\end{figure}

In this section, we analyze the effect of various components of \method{} on its performance. For this, we evaluate various versions of our model with cumulatively removed components. The experimental results are presented in \reffig{fig:ablation}. We observe that on removing different components from \method{}, the performance of the model degrades drastically. The results validate that GCNs are effective at encoding syntactic information. Further, the improvement from side information shows that it is complementary to the features extracted from text, thus validating the central thesis of this paper, that inducing side information leads to improved relation extraction. 

\subsection{Effect of Relation Alias Side Information}
\label{sec:results_rel_side}
In this section, we test the performance of the model in setting where relation alias information is not readily available. For this, we evaluate the performance of the model on four different settings:

\begin{itemize}[itemsep=2pt,topsep=2pt,parsep=0pt,partopsep=0pt,leftmargin=*]
	\item \textbf{None:} Relation aliases are not available.
	\item \textbf{One:} The name of relation is used as its alias. 
	\item \textbf{One+PPDB:} Relation name extended using Paraphrase Database (PPDB). 
	\item \textbf{All:} Relation aliases from Knowledge Base\footnote{Each relation in Riedel dataset is manually mapped to corresponding Wikidata property for getting relation aliases. Few examples are presented in supplementary material.}

\end{itemize}

\begin{figure}[t]
	\centering
	\includegraphics[width=\columnwidth]{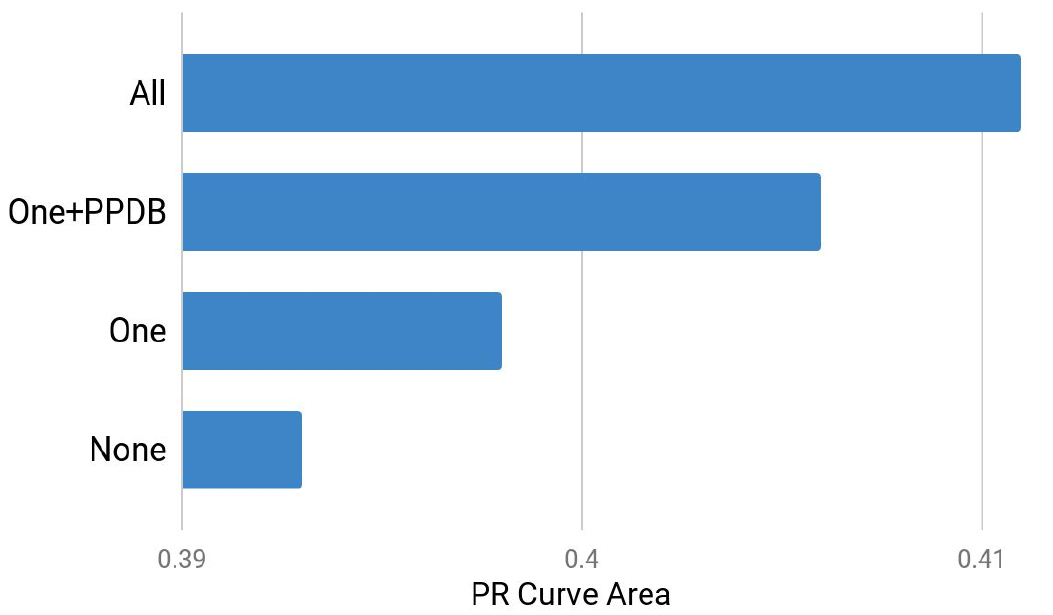}
	\caption{\label{fig:diff_aliases} Performance on settings defined in \refsec{sec:results_rel_side} with respect to the presence of relation alias side information on Riedel dataset. \method{} performs comparably in the absence of relations from KB.}
\end{figure}

The overall results are summarized in \reffig{fig:diff_aliases}. We find that the model performs best when aliases are provided by the KB itself. Overall, we find that \method{} gives competitive performance even when very limited amount of relation alias information is available. We observe that performance improves further with the availability of more alias information.

\section{Conclusion}
\label{sec:conclusion}

In this paper, we propose \method{}, a novel neural network based model which makes principled use of relevant side information, such as entity type and relation alias, from Knowledge Base, for improving distant supervised relation extraction. \method{} employs Graph Convolution Networks for encoding syntactic information of sentences and is robust to limited side information. Through extensive experiments on benchmark datasets, we demonstrate \method{}'s effectiveness over state-of-the-art baselines. We have made \method{}'s source code publicly available to promote reproducible research.

\section*{Acknowledgements}
We thank the anonymous reviewers for their constructive comments. This work is supported in part by the Ministry of Human Resource Development (Government of India) and by a gift from Google.

\bibliography{emnlp2018}
\bibliographystyle{acl_natbib_nourl}

\end{document}